\documentclass[10pt, letterpaper, journal, twocolumn, final]{IEEEtran}
\usepackage{amsmath}
\usepackage[hidelinks,hyperindex,breaklinks]{hyperref}
\usepackage{amssymb}
\usepackage{booktabs}
\usepackage{cite}
\usepackage{graphicx}
\usepackage{setspace}
\allowdisplaybreaks

\renewcommand{\vec}[1]{\mathnormal{#1}}
\renewcommand{\leq}{\leqslant}
\renewcommand{\geq}{\geqslant}

\DeclareMathOperator*{\tr}{tr}

\title{Enhanced Low-Rank Matrix Approximation}
\author{
{Ankit Parekh and Ivan W. Selesnick \vspace{-1em}}
\thanks{Copyright (c) 2015 IEEE. Personal use of this material is permitted. However, permission to use this material for any other purposes must be obtained from the IEEE by sending a request to pubs-permissions@ieee.org.}
\thanks{A. Parekh (ankit.parekh@nyu.edu) is with the Department of Mathematics, Tandon School of Engineering, New York University, and I. Selesnick is with the Department of Electrical and Computer Engineering, Tandon School of Engineering, New York University.} \thanks{This work was supported by the ONR under Grant No. N00014-15-1-2314 and by the NSF under Grant No. CCF-1525398.}
\thanks{MATLAB software is available at \url{htpps://goo.gl/xAi85N} on IEEE Xplore.}
}

\begin{document}
\maketitle

\begin{abstract}
This letter proposes to estimate low-rank matrices by formulating a convex optimization problem with non-convex regularization. We employ parameterized non-convex penalty functions to estimate the non-zero singular values more accurately than the nuclear norm. A closed-form solution for the global optimum of the proposed objective function (sum of data fidelity and the non-convex regularizer) is also derived. The solution reduces to singular value thresholding method as a special case. The proposed method is demonstrated for image denoising. 
\end{abstract}
\begin{IEEEkeywords}
Low-rank matrix, nuclear norm, image denoising, convex optimization, non-convex regularization.
\end{IEEEkeywords}

\section{Introduction}

Approximating a given matrix by a low-rank matrix is a fundamental problem in many signal processing applications \cite{Jeong2014, Markovsky2008, Achlioptas2001, Hansson2012, Fazel2013,Nguyen2013, Nadakuditi2014}. The low-rank matrix approximation (LRMA) problem is a pivotal step in numerous machine learning \cite{Kannan2008, Drineas2006,Foygel2012,Foygel2011}, statistical signal processing  \cite{Cadzow1988,Vempala2002,Sanjeev2001,Dasgupta1999, Tipping1999, Richard2012}, graph signal processing \cite{Chatterjee2012}, and tensor recovery \cite{Huang2015} problems. We consider the problem of estimating a low-rank matrix $X$ from its noisy observation $Y$, 
\begin{align}
\label{eq::signal model}
Y = X + W, \quad X,Y,W \in \mathbb{R}^{m \times n}
\end{align} where $W$ represents zero-mean additive white Gaussian noise (AWGN). We define the LRMA problem as
\begin{align}
\label{eq::cost Function NNM}
\arg\min_{X} \left\lbrace \Psi(X) = \frac{1}{2}\|Y-X\|_F^2 + \lambda\sum_{i=1}^{k}\phi(\sigma_i(X);a)\right\rbrace,
\end{align} where $k = \min(m,n)$, $\sigma_i(X)$ is the i$^\text{th}$ singular value of the matrix $X$, and $\phi$ is a sparsity-inducing regularizer, possibly non-convex. The standard nuclear norm minimization (NNM) problem \cite{Cai2008} is a special case of the LRMA problem \eqref{eq::cost Function NNM}, with $\phi(x) = |x|$. Note that the NNM problem is convex and its global minimum can be directly obtained using the singular value decomposition (SVD) of the input matrix $Y$. In particular, if $Y = U \Sigma V^T$ is the SVD of $Y$, then the solution to the NNM problem is given by
\begin{align}
\label{SVT}
\hat{X} = U \cdot \mbox{soft(}\Sigma;\lambda) \cdot V^T
\end{align} where $\mbox{soft(}\cdot ; \cdot)$ is the soft-threshold function \cite{Donoho1995} applied to the singular values of $Y$. The solution \eqref{SVT} to the NNM problem is known as `Singular Value Thresholding' (SVT) method \cite{Cai2008}. 

The SVT method tends to underestimate the non-zero singular values. Several recent studies have emphasized the benefit of non-convex penalty functions compared to the nuclear norm for the estimation of singular values \cite{Lu2014,Lu2014b,Wang2013,Chartrand2012}. However, the use of non-convex penalty functions generally suffers from numerous issues (spurious local minima, initialization issues, etc.).

In this letter, we aim to estimate the non-zero singular values more accurately than the nuclear norm, while maintaining convexity of the objective function. To this end, we propose to use a particular class of parameterized non-convex penalty functions. We show how to set the non-convex penalty parameter to ensure the proposed objective function \eqref{eq::cost Function NNM} is strictly convex. The idea of using non-convex penalty functions within a convex optimization framework was described by Blake and Zisserman \cite{Blake1987} and Nikolova \cite{Nikolova1998}, and has been applied to various signal processing problems \cite{Selesnick2015,Parekh2015,Parekh2015b,Lanza2015,He2016}.

\subsection{Related Work}
\label{subsec::Related Work}
Several non-convex penalty functions have been utilized for the LRMA problem \eqref{eq::cost Function NNM}: the weighted nuclear norm \cite{Gu2014}, transformed Schatten-1 (TS1) \cite{Zhang2015} and the proximal p-norm \cite{Chartrand2012}. The use of these non-convex penalty functions makes the overall LRMA problem non-convex. As such, iterative algorithms aiming to reach a stationary point (i.e., not necessarily global optimum) of the non-convex objective function have been developed \cite{Zhang2015,Gu2014}. Also, a non-iterative locally optimal solution for the LRMA problem using the proximal p-norm is reported in \cite{Chartrand2012}. Note that the proximal operators (see Sec.~\ref{subsection::Penalty Functions}) associated with the TS1 penalty and the proximal p-norm are not continuous for all values of the regularization parameter $\lambda$. In contrast, the proposed approach always leads to a convex problem formulation. A broader class of non-convex penalty functions was studied for the LRMA problem \eqref{eq::cost Function NNM} in \cite{Wang2013,Lu2014b,Lu2016}. The iteratively reweighted nuclear norm minimization \cite{Lu2014,Lu2016} and generalized singular value thresholding \cite{Lu2014b} methods provide a locally optimal solution to the LRMA problem, provided that the penalty functions satisfy certain assumptions (see Assumption A1 of \cite{Lu2014}). 

\section{Preliminaries}
We denote vectors and matrices by lower and upper case letter respectively. The Frobenius norm of a matrix $\vec{Y} \in \mathbb{R}^{m \times n}$ is defined as
\begin{align}
\label{eq::Entry-wise Norms}
\|\vec{Y}\|_F^2 &= \sum_{i=1}^{m}\sum_{j=1}^{n} |\vec{Y}_{i,j}|^2 \\
				&= \tr(Y^T Y).
\end{align} The nuclear norm of $Y$ is defined as
\begin{align}
\label{eq::Nuclear Norm}
\|\vec{Y}\|_* = \sum_{i=1}^{\min(m,n)} \sigma_i(\vec{Y}), 
\end{align} where $\sigma_i(\vec{Y})$ represents the $i^{\text{th}}$ singular value of $\vec{Y}$. 

\subsection{Penalty Functions, Proximity Operators}
\label{subsection::Penalty Functions}
In order to estimate non-zero singular values more accurately, and induce sparsity more effectively, than the nuclear norm, we use non-convex penalty functions parameterized by the parameter $a \geq  0$ \cite{Selesnick_2014_MSC}. We make the following assumption on such penalty functions.
\newtheorem{assumption}{\bf Assumption}
\begin{assumption}
\label{theorem::assumption 1}
The non-convex penalty function $\phi\colon\mathbb{R} \to \mathbb{R}$ satisfies the following properties
\begin{enumerate}
\item $\phi$ is continuous on $\mathbb{R}$, $\phi$ is continuously differentiable on $\mathbb{R}\setminus\lbrace 0 \rbrace$ and symmetric, i.e., $\phi(-x; a) = \phi(x; a)$
\item $\phi'(x;a) \geq 0, \forall x > 0$	
\item $\phi''(x;a) \leq 0, \forall x > 0$
\item $\phi'(0^{+};a) = 1$
\item $\inf\limits_{x\neq0}\phi''(x;a) = \phi''(0^+;a) = -a$
\end{enumerate}
\end{assumption} An example of a penalty function $\phi$ satisfying Assumption 1 is the partly quadratic penalty function \cite{Zhang2010,Bayram2015} defined as
\begin{align}
\label{eq::rat penalty}
\phi(x;a) := \begin{cases}
|x|-\dfrac{a}{2}x^2, & |x| \leqslant \frac{1}{a} \\
\dfrac{1}{2a}, & |x| \geqslant \frac{1}{a}
	\end{cases}
\end{align}Note that as $a \to 0$, the $\ell_1$ norm is recovered as a special case of this penalty. Several other examples of penalty functions satisfying Assumption \ref{theorem::assumption 1} are listed in Table~II of Ref.~\cite{Chen2014}.

\newtheorem{definition}{\bf Definition}
\begin{definition}{\cite{Combettes2007,Combettes2011}.}
	\label{def::prox}
	Let $\phi\colon\mathbb{R}\to\mathbb{R}$ be a non-convex penalty function satisfying Assumption 1. The proximal operator of $\phi$, $\Theta:\mathbb{R}\to\mathbb{R}$, is defined as
	\begin{align}
	\label{theta}
		\Theta(y; \lambda,a) := \arg\min_{x \in \mathbb{R}} \left\lbrace \frac{1}{2}(y-x)^2 + \lambda\phi(x;a) \right\rbrace.
	\end{align} If $0 \leqslant a < 1/\lambda$, then $\Theta$ is a continuous non-linear threshold function with threshold value $\lambda$, i.e., 
	\begin{align}
	\Theta(y; \lambda,a) = 0, \quad  \forall |y| < \lambda.
	\end{align}
\end{definition} For example, the proximal operator of the partly quadratic penalty \eqref{eq::rat penalty} is the firm threshold function \cite{Gao1997} defined as
\begin{align*}
\Theta(y; \lambda,a) := \min \bigl\lbrace |y|, \max\lbrace (|y| - \lambda)/(1-a\lambda), 0 \rbrace\bigr\rbrace\text{sign}(y).
\end{align*} Note that for the matrix $\vec{X}$, the notation $\Theta(X; \lambda,a)$ indicates that the proximal operator is applied element-wise to $\vec{X}$. 

\section{Low-rank matrix approximation}
\label{section::LRA}

\subsection{Convexity condition}
We note the following lemmas, which will be used to obtain a convexity condition for the objective function $\Psi$ in \eqref{eq::cost Function NNM}.
\newtheorem{lemma}{\bf Lemma}
\begin{lemma}{\cite{Parekh2015}.}
\label{lemma 1}
Let $\phi\colon\mathbb{R}\to\mathbb{R}$ be a non-convex penalty function satisfying Assumption \ref{theorem::assumption 1}. The function $s\colon\mathbb{R} \to \mathbb{R}$ defined as
\begin{align}
\label{eq::s function}
s(x;a) := \phi(x;a) - |x|,
\end{align} is continuously differentiable, concave, and satisfies
\begin{align}
\label{eq::lemma 1 inequality}
-a \leqslant s''(x;a) \leqslant 0. 
\end{align}
\end{lemma}

\begin{lemma}
	\label{lemma 2}
	Let $\phi\colon\mathbb{R}\to\mathbb{R}$ be a non-convex penalty function satisfying Assumption 1 and let $\lambda > 0$. The function $f\colon\mathbb{R}\to\mathbb{R}$ defined as
	\begin{align}
	\label{eq::scalar function f}
	f(x) := \dfrac{1}{2}x^2 + \lambda\phi(x;a)
	\end{align} is strictly convex if
	\begin{align}
	\label{eq::condition for scalar function f}
	0 \leqslant a < \dfrac{1}{\lambda}.
	\end{align}
\end{lemma}

\begin{IEEEproof}
Consider the function $g\colon\mathbb{R}\to\mathbb{R}$ defined as 
\begin{align}
	\label{eq::scalar g}
	g(x) := \dfrac{1}{2} x^2 + \lambda s(x;a),
\end{align}	where the function $s$ is defined in \eqref{eq::s function}. Note that 
\begin{align}
	\label{eq::connection between f and g}
	f(x) &= \dfrac{1}{2}x^2 + \lambda \phi(x;a) \\
	     &= \dfrac{1}{2}x^2 + \lambda \bigl (s(x;a) + |x| \bigr) \\
	     &= g(x) + \lambda |x|.
\end{align} 

In order to ensure the strict convexity of the function $f$, it is sufficient to ensure the strict convexity of the function $g$. To this end, it suffices to show that the second derivative of the function $g$ is positive, i.e. $g''(x) > 0$ for all $x \in \mathbb{R}$. From \eqref{eq::scalar g}, we note that $g''(x) > 0$ if
\begin{align}
\label{eq::second derivative of g}
1 + \lambda s''(x;a) > 0. 
\end{align}
Using Lemma 1, $g''(x) > 0$ for all $x \in \mathbb{R}$ if $0 \leqslant a < 1/\lambda$. Thus, the function $g$ (and hence the function $f$) is strictly convex if the parameter $a$ satisfies the inequality \eqref{eq::condition for scalar function f}. 
\end{IEEEproof}

The following theorem shows $a$ must satisfy \eqref{eq::condition for scalar function f} to ensure the strict convexity of the function $\Psi$ in \eqref{eq::cost Function NNM}. 
\newtheorem{theorem}{\bf Theorem}
\begin{theorem}
\label{theorem::Theorem 1}
Let $\phi\colon\mathbb{R}\to\mathbb{R}$ be a non-convex penalty function satisfying Assumption \ref{theorem::assumption 1}. The objective function $\Psi\colon\mathbb{R}^{m \times n}\to\mathbb{R}$ in \eqref{eq::cost Function NNM} is strictly convex if $0 \leq a < 1/\lambda.$
\end{theorem}

\begin{IEEEproof}
We assume $m = n$. The derivation for $m \neq n$ is similar. Let $s\colon\mathbb{R} \to \mathbb{R}$ be defined as in Lemma~1. Consider the function $G\colon\mathbb{R}^{m \times n} \to \mathbb{R}$ defined as
\begin{align}
\label{eq:: Psi hat}
G(X) &= \dfrac{1}{2}\|Y-X\|_F^2 + \lambda\sum_{i=1}^{m}s\bigl(\sigma_i(X); a \bigr) \\
	 &= \dfrac{1}{2}\tr\bigl( (Y-X)^T(Y-X) \bigr) + \lambda\sum_{i=1}^{m}s\bigl(\sigma_i(X); a \bigr) \nonumber \\
	 &= \dfrac{1}{2}\tr(Y^TY) - \tr(XY^T) + \dfrac{1}{2}\tr(X^TX) \nonumber \\
	 & \qquad + \lambda\sum_{i=1}^{m}s\bigl(\sigma_i(X); a\bigr).
\end{align} Note that $\tr(Y^TY)$ does not depend on $X$ and $\tr(XY^T)$ is linear in $X$. Hence, the function $G$ is strictly convex if $G_2$ is strictly convex, where $G_2 \colon\mathbb{R}^{m \times n} \to \mathbb{R}$ is defined as
\begin{align}
\label{eq::sigmaX}
G_2(X) &= \dfrac{1}{2} \tr(X^TX) + \lambda \sum_{i=1}^{m} s\bigl( \sigma_i(X);a \bigr) \\
\label{eq::G_2 final form}
       &= \dfrac{1}{2} \sum_{i=1}^{m} \sigma_i^2(X) + \lambda \sum_{i=1}^{m} s\bigl( \sigma_i(X);a \bigr). 
\end{align} 

To ensure that the function $G_2$ is strictly convex, consider the function $h\colon\mathbb{R}^m\to\mathbb{R}$ defined as
\begin{align}
	\label{eq::definition of function h}
	h(x) = \sum_{i=1}^{m} g(x_i),
\end{align}where $g$ is defined as in \eqref{eq::scalar g}. Due to Lemma 2, the function $h$ is strictly convex and absolutely symmetric \cite{Lewis1995}. In view of \eqref{eq::scalar g}, \eqref{eq::G_2 final form}, and \eqref{eq::definition of function h}, the function $G_2$ can be written as
\begin{align}
	\label{eq::G_2}
	G_2(X) &= \dfrac{1}{2} \sum_{i=1}^{m} \sigma_i^2(X) + \lambda\sum_{i=1}^{m} s\bigl( \sigma_i(X);a \bigr) \\
		   &= \sum_{i=1}^{m} \biggl( \dfrac{1}{2}\sigma_i^2(X) + \lambda s \bigl( \sigma_i(X);a \bigr) \biggr)\\
		   &= \sum_{i=1}^{m} g(\sigma_i(X)) \\
		   &= h(\sigma(X)),
\end{align} where $\sigma(X)$ denotes the vector of singular values of the matrix $X$. From Corollary 2.6 of \cite{Lewis1995}, since $h$ is absolutely symmetric, the unitarily invariant function $G_2(X) = h(\sigma(X))$ is strictly convex if and only if $h$ is strictly convex. It follows from Lemma 2 that $G_2$ (and hence $G$) is strictly convex if $a$ satisfies inequality \eqref{eq::condition for scalar function f}. 

Further, note that $\Psi$ in \eqref{eq::cost Function NNM} can be written as
\begin{align}
\Psi(X) &:= \frac{1}{2}\|Y-X\|_F^2 + \lambda\sum_{i=1}^{m}\phi(\sigma_i(X); a) \\
&= \frac{1}{2}\|Y-X\|_F^2 + \lambda\sum_{i=1}^{m}\Bigl( s(\sigma_i(X); a) + |\sigma_i(X)| \Bigr) \nonumber \\
&= \frac{1}{2}\|Y-X\|_F^2 + \lambda\sum_{i=1}^{m}s(\sigma_i(X);a) + \lambda\sum_{i=1}^{m}|\sigma_i(X)| \nonumber \\
&= G(X) + \lambda\|X\|_*. 
\end{align} Hence, if $a$ satisfies \eqref{eq::condition for scalar function f}, then the function $\Psi$ in \eqref{eq::cost Function NNM} is strictly convex, being the sum of a convex function (the nuclear norm) and the strictly convex function $G$.  
\end{IEEEproof}

\subsection{Global optimum}
Theorem \ref{theorem::Theorem 1} ensures the proposed objective function $\Psi$ in the LRMA problem \eqref{eq::cost Function NNM} is strictly convex if $ 0 \leqslant a < 1/\lambda$. The following theorem provides a closed-form solution to the proposed LRMA problem \eqref{eq::cost Function NNM}. The solution involves thresholding the singular values of the input matrix $Y$.
\begin{theorem}
Let $Y = U \Sigma V^T$ be the SVD of $Y$ and $\phi\colon\mathbb{R}\to\mathbb{R}$ be a non-convex penalty function satisfying Assumption 1. If $0 \leqslant a < 1/\lambda$, then the global minimizer of \eqref{eq::cost Function NNM} is
\begin{align}
\label{eq::solution}
\bar{X} = U\cdot \Theta(\Sigma; \lambda,a)\cdot V^T,
\end{align} where $\Theta$, defined in \eqref{theta}, is the threshold function associated with the non-convex penalty function $\phi$. 
\end{theorem}
\begin{IEEEproof}
We let $m = n$. The proof for $m \neq n$ is similar. Since $0 \leqslant a < 1/\lambda$, the function $\Psi$ in \eqref{eq::cost Function NNM} is strictly convex; hence the minimizer of \eqref{eq::cost Function NNM} is unique. Let 
\begin{align}
\Phi(X) := \sum_{i=1}^{m} \phi\bigl(\sigma_i(X);a\bigr).
\end{align} Note that for unitary matrices $U$ and $V$, $\Phi(X) = \Phi(UXV)$. Using the unitary invariant property of the Frobenius norm, and the SVD of $Y$, we write
\begin{align}
\hat{X} &= \arg\min_{X} \left\lbrace \frac{1}{2}\|Y-X\|_F^2 + \lambda\Phi(X) \right\rbrace \\
&=\arg\min_{X} \left\lbrace \frac{1}{2}\|\Sigma-U^T X V\|_F^2 + \lambda \Phi(U^T X V) \right\rbrace \\
\label{eq::diagonal form of cost function}
&= U \cdot \arg\min_{X} \left\lbrace \frac{1}{2}\|\Sigma - X\|_F^2 + \lambda \Phi(X) \right\rbrace \cdot V^T.
\end{align} As a result of \eqref{eq::diagonal form of cost function}, we need to show that 
\begin{align}
\label{eq:: diagonal cost function}
\Theta(\Sigma; \lambda,a) := \arg\min_{X} \left\lbrace \frac{1}{2}\|\Sigma - X\|_F^2 + \lambda \Phi(X) \right\rbrace, 
\end{align} is the optimal solution. Note that with $Y = \Sigma$, \eqref{eq:: diagonal cost function} is strictly convex (as $0 \leqslant a < 1/\lambda$) from Theorem \ref{theorem::Theorem 1} and hence admits a unique global minimum. Let  
\begin{align}
X = U_x \Sigma_x V_x^T, 
\end{align} be the SVD of $X$. We can write
\begin{align}
\|\Sigma - X\|_F^2 &= \|X\|_F^2 + \|\Sigma\|_F^2 - 2 \tr(X^T \Sigma) \\
\label{eq:: von neumann trace}
&\geqslant \|\Sigma_x\|_F^2 + \|\Sigma\|_F^2 - 2\tr (\Sigma_x\Sigma) \\
\label{eq::theorem 2 inequality}
&= \|\Sigma - \Sigma_x\|_F^2.
\end{align} The inequality in \eqref{eq:: von neumann trace} is due to von Neumann's trace inequality. From \eqref{eq::theorem 2 inequality}, we note that
\begin{align}
\label{eq::equality of diagonal}
\frac{1}{2}\|\Sigma - X\|_F^2 + \lambda \Phi(X) \geqslant \frac{1}{2}\|\Sigma - \Sigma_x\|_F^2 + \lambda \Phi(\Sigma_x),
\end{align} with equality if $X = \Sigma_x$. Note that $\Sigma_x$ is a diagonal matrix. 

Consider the problem of finding such a diagonal matrix $\Sigma_x$ using the optimization problem
\begin{align}
\label{eq::Sigma cost function}
\arg\min_{\Sigma_x} \left\lbrace \frac{1}{2}\|\Sigma - \Sigma_x\|_F^2 + \lambda\Phi(\Sigma_x)\right\rbrace.
\end{align} The optimization problem \eqref{eq::Sigma cost function} is separable, as $\Sigma$ and $\Sigma_x$ are diagonal. Hence, the solution to \eqref{eq::Sigma cost function} can be obtained by applying the threshold function $\Theta$ to the entries of $\Sigma$. Thus, the optimal solution to \eqref{eq::Sigma cost function} is 
\begin{align}
\Theta(\Sigma; \lambda,a) = \arg\min_{\Sigma_x} \left\lbrace \frac{1}{2}\|\Sigma - \Sigma_x\|_F^2 + \lambda\Phi(\Sigma_x)\right\rbrace.
\end{align}Using $X = \Theta(\Sigma; \lambda,a)$, we obtain the equality in \eqref{eq::equality of diagonal}. This completes the proof of \eqref{eq:: diagonal cost function}.
\end{IEEEproof} Note that due to the monotonocity of the threshold function, the solution \eqref{eq::solution} to the proposed LRMA problem \eqref{eq::cost Function NNM} does not change the order of singular values.

\section{Examples}
\label{sec::Examples}
\subsection{Synthetic data}
We apply the proposed enhanced low-rank matrix approximation (ELMA) method on synthetic data to assess its performance. For the following example, we generate 15 realizations of a random  matrix $M \in \mathbb{R}^{m \times n}$ with rank $k$, such that 
\begin{align}
\label{eq::creating random matrix}
M := AB, \quad A \in \mathbb{R}^{m \times k}, B \in \mathbb{R}^{k \times n}.
\end{align} The matrices $A$ and $B$ are random matrices with entries chosen from an i.i.d normal distribution. We let $M$ be a $200 \times 200$ matrix with rank 100. We add white Gaussian noise ($1 \leqslant \sigma \leqslant 10$) to the matrix $M$, thus creating the noisy observation matrix $Y$. We use the normalized root square error (RSE) defined as $\mbox{RSE} = \|X-M\|_F/ \|M\|_F$, as a performance measure.
\begin{figure}
	\centering
	\includegraphics[scale = 0.9]{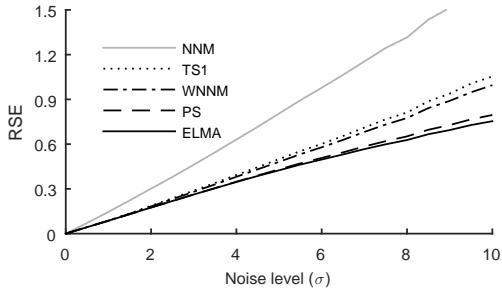}
	\caption{Average RSE as a function of the noise level ($\sigma$).}
	\label{fig::Average RSE}
\end{figure}

We compare the proposed ELMA method to the weighted nuclear norm minimization (WNNM) \cite{Gu2014}, standard nuclear norm minimization (NNM) \cite{Cai2008}, p-shrinkage (PS) \cite{Chartrand2012} and the TS1 \cite{Zhang2015} LRMA methods. For the ELMA, NNM and PS methods, we set $\lambda = \beta \sigma$, where $\beta$ is manually set to optimize the RSE for each method. We use ELMA with the partly quadratic penalty \eqref{eq::rat penalty} with $a = 0.6/\lambda$ (i.e., we use the firm threshold function \cite{Gao1997}). For the PS method, we use $p = -2$. For the WNNM method, we set the weights inversely proportional to the singular values of the input matrix, as suggested in \cite{Gu2014}. Figure \ref{fig::Average RSE} shows the average RSE for the different methods. The proposed ELMA method yields the lowest RSE values for most values of $\sigma$.

\subsection{Image Denoising}
We consider denoising an image $X$ from its noisy observation $Y$, where the noise is AWGN with noise level $\sigma$. Recently, a growing number of image denoising methods are emerging based on the non-local self similarity (NSS) approach \cite{Buades2005,Yang2013,Manjon2008, Ji2010}. State of the art image denoising algorithms such as BM3D \cite{Dabov2007}, LSSC\cite{Mairal2009}, NCSR\cite{Dong2011}, and PLR \cite{Hu2015} are NSS based methods. 

For a local patch in a noisy image $Y$, denote the matrix formed by stacking non-local similar patches as $Y_j$. The non-local similar patches can be found using block matching methods \cite{Dabov2007}. As such, we write $Y_j = X_j + W_j$, where $X_j$ is the patch matrix of the clean image and $W_j$ is AWGN. In order to estimate $X_j$ from $Y_j$, we propose the following objective function, similar to Eq. (9) of \cite{Gu2014},
\begin{align}
\label{eq::Image cost function}
\hat{X}_j := \arg\min_X \left\lbrace \frac{1}{2}\|Y_j - X\|_F^2 + \lambda\sum_{i=1}^{m}\phi(\sigma_i(X);a)\right\rbrace.
\end{align} 

The objective function in \eqref{eq::Image cost function} is strictly convex if $a < 1/\lambda$, from Theorem 1. Further, with $Y_j = U_j \Sigma_j V_j^T$ as the SVD of $Y_j$, the solution to \eqref{eq::Image cost function} is given by $\hat{X}_j = U_j \Theta(\Sigma_j; \lambda,a) V_j^T$, as per Theorem 2. 

We perform image denoising using ELMA to estimate each patch matrix $X_j$. We compare the denoised images obtained using ELMA with the denoised images obtained using BM3D, PS, NNM and the WNNM methods. We use three test images of size $512 \times 512$ and add AWGN with $\sigma = 100$. We set the regularization parameter $\lambda$ in \eqref{eq::Image cost function} as in the previous example. 

\begin{figure}
	\centering
	\includegraphics[scale = 0.8]{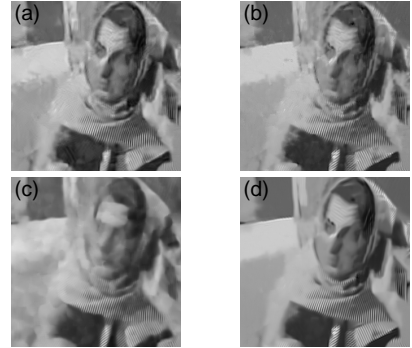}
	\caption{Denoised `Barbara' image using several LRMA methods within the NSS framework. (a) BM3D (b) WNNM (c) p-shrinkage (d) ELMA (proposed).}
	\label{fig::comparison of image denoising}
\end{figure}

\begin{table}
\centering
\caption{Average PSNR (dB) obtained by several image denoising methods.}
\begin{tabular}{@{}lccccc@{}}
\toprule 
Image & \multicolumn{5}{c}{Method} \\
\cmidrule(l){2-6}
 & BM3D & PS & NNM & WNNM & ELMA \\
\midrule
Boat & 23.97 & 23.82 & 22.88 & 24.10 & 24.03\\
Barbara & 23.62 & 23.55 & 22.98 & 24.37 & 24.46\\
Couple & 23.51 & 23.35 & 22.07 & 23.57 & 23.53\\
\bottomrule
\end{tabular}
\label{table::PSNR}	
\end{table}

Figure \ref{fig::comparison of image denoising} shows the denoised `Barbara' image using several LRMA methods. The average PSNR values for the three test images are listed in Table~\ref{table::PSNR}. The proposed ELMA method achieves higher PSNR than the BM3D, PS, and NNM methods. On average, the PSNR values obtained by the ELMA method are comparable to those obtained by the WNNM method. However, in terms of the image quality, the proposed ELMA method contains fewer artifacts.

\section{Conclusion}

This letter proposes a method to estimate low-rank matrices corrupted with AWGN. The proposed method, which outperforms nuclear norm minimization, is based on an objective function comprising a data-fidelity term and a non-convex regularizer. The proposed objective function is convex even though the regularizer is not. The non-iterative solution to the proposed objective function is obtained by thresholding the singular values of the input matrix. We demonstrate the effectiveness of the proposed method for image denoising by using it within a non-local self-similarity based image denoising algorithm. The proposed method has the same computational complexity as the SVT method. As such, scalable methods \cite{Shang2016} to accelerate the computation may also be applicable to the proposed method.

\section*{Supplemental Figures}

\begin{figure}[h]
	\centering
	\includegraphics{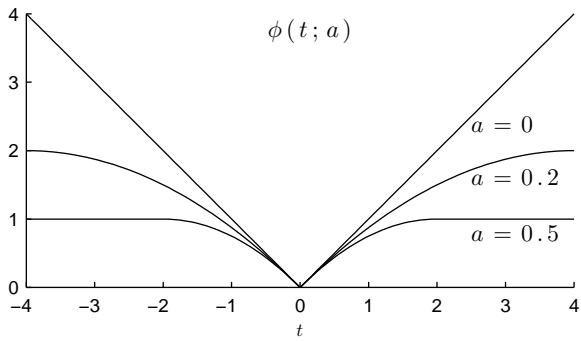}
	\caption{Penalty function $ \phi( t ; a) $ for several values of $ a $.
	}
	\label{fig:pen_vara}
\end{figure}

\begin{figure}[h]
	\centering
	\includegraphics{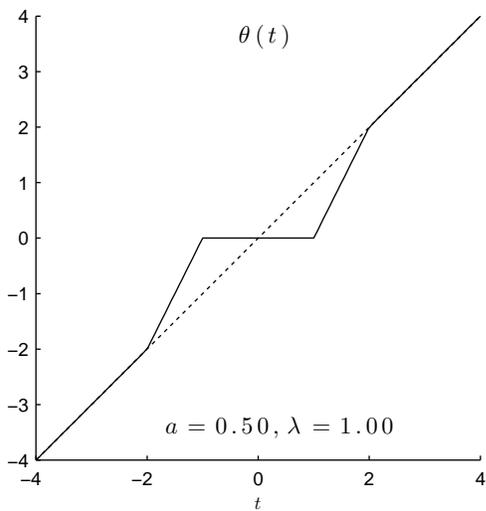}
	\caption{Firm threshold function.}
	\label{fig:thresh}
\end{figure}

Figure \ref{fig:pen_vara} illustrates the particular penalty function $ \phi(t; a) $ defined in (7) for 
several values of the parameter $ a $.
Figure~\ref{fig:thresh} illustrates the threshold function (i.e., proximity operator) associated
with this penalty. This is known as the \emph{firm} threshold function. 

\begin{figure}
	\centering
	\includegraphics{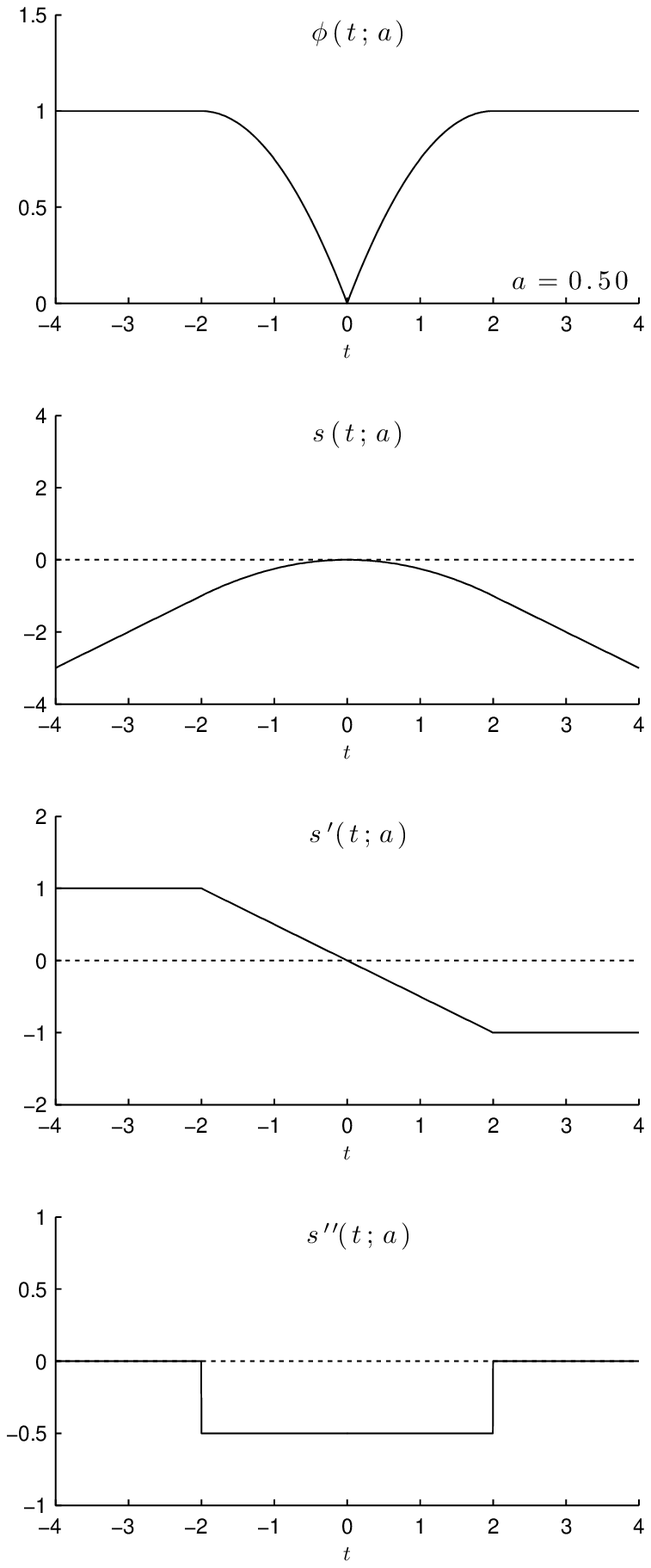}
	\caption{Penalty $ \phi( t ; a) $ and function $ s( t ; a ) = \phi(t; a) - |t| $ for $ a = 0.5 $.}
	\label{fig:pen_pqp}
\end{figure}

Figure \ref{fig:pen_pqp} illustrates the penalty function and the corresponding function $ s(t; a) $
defined in (10).
The figure also shows the first and second-order derivatives of the function $ s(t ; a) $.

\vfill
\vfill

\vfill

\end{document}